\documentclass[conference]{IEEEtran}
\let\labelindent\relax
\usepackage{times}
\usepackage[T1]{fontenc}
\let\labelindent\relax
\usepackage{enumitem}
\usepackage{amsmath}
\usepackage{amssymb}
\usepackage{graphicx}

\usepackage[numbers]{natbib}
\usepackage{multicol}
\usepackage[bookmarks=true]{hyperref}
\usepackage{ulem}
\usepackage{color}

\usepackage{tikz}
\newcommand*\circled[1]{\tikz[baseline=(char.base)]{
            \node[shape=circle,draw,inner sep=1pt, scale=0.8] (char) {#1};}}

\newcommand{\edit}[2]{#2}

\makeatletter
\def\blfootnote{\gdef\@thefnmark{}\@footnotetext}
\makeatother

\begin{document}

\title{Computational Design of a Low-Visibility UAV Using a Human-Aligned Perceptual Metric}



%

\author{
    Jingxian Wang$^{1*}$, 
    Chen Yu$^{1*}$, 
    David Matthews$^{1}$, 
    Emma Alexander$^{1}$, 
    Sam Kriegman$^{1}$, 
    Michael Rubenstein$^{1\dagger}$
}

\maketitle

\blfootnote{$^*$Contributed equally to the paper.}
\blfootnote{$^1$Northwestern University, Evanston, Illinois, USA}
\blfootnote{$^\dagger$Corresponding author. E-mail: rubenstein@northwestern.edu}

\begin{abstract}
We introduce Phantom Twist, a type of single-propeller UAV designed to achieve low visibility through high-speed spinning and the exploitation of motion blur. We develop a two-stage automated design pipeline that optimizes the placement of functional components including batteries, control PCB, motor-propeller assembly, and counterweights. The pipeline minimizes visibility as measured by a human-aligned perceptual metric (LPIPS) while strictly satisfying inertial and aerodynamic constraints required for stable flight. We validate this approach through fabrication and flight testing of multiple prototypes. These tests confirm that our pipeline produces stable, controllable designs and that the optimized UAV exhibits significantly reduced visual perceptibility compared to conventional quadcopters.
\end{abstract}

\IEEEpeerreviewmaketitle

\section{Introduction}

In a wide variety of settings, awareness of a robot’s presence can negatively impact its effectiveness or the validity of the data it collects. Robots deployed for inspection, environmental monitoring, or human–robot interaction may distract, disturb, or otherwise alter the behavior of nearby humans, animals, or other systems simply through their sensed presence. These effects can bias observations, reduce task performance, and limit deployment in sensitive environments. Minimizing the influence of a robot on its surroundings enables closer observation of natural behavior, more reliable interaction, and broader applicability of robotic systems in real-world settings.

To mitigate these effects, prior research has explored various hardware designs and behavioral strategies intended to lower a robot's perceptibility. Traditionally, these efforts focus on reducing visual salience through static or active coloration. Drawing from biological camouflage found in plants and animals~\cite{stevens2008camouflage}, some systems adapt their surface colors and patterns to blend into the environment~\cite{li2018distributed, morin2012camouflage, kim2021biomimetic}. This logic extends to active color emitters~\cite{palovuori2021toward} and counter-illumination: a technique exemplified by the ``Yehudi light''~\cite{ndrc1946camouflage} as early as the 1940s. More recently, researchers have also explored material transparency, using clear actuators and sensors to make robots' body optically permeable~\cite{li2019transparent,won2021transparent}.


\begin{figure}[t]
    \centering
    \includegraphics[width=\linewidth]{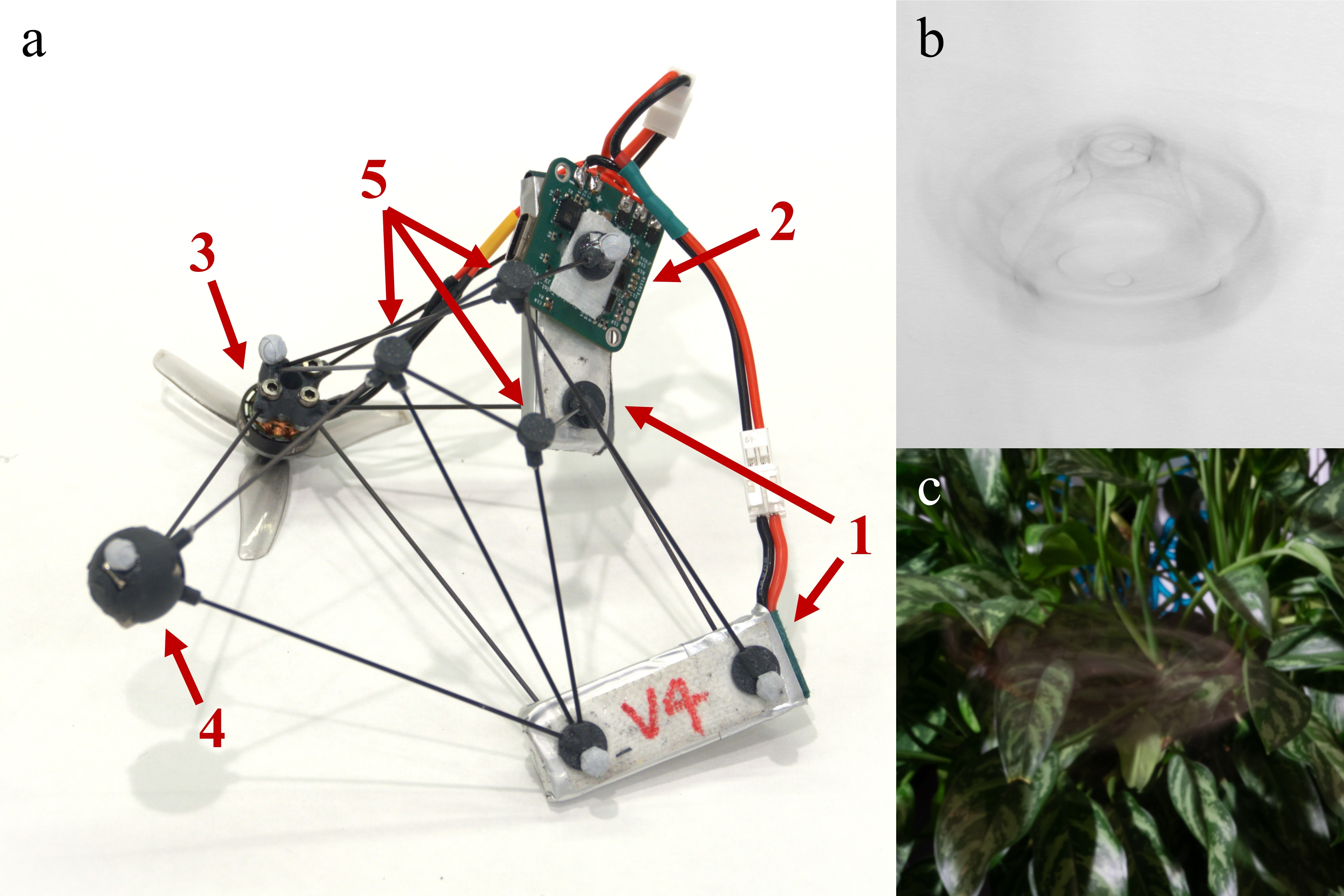}
    \caption{\textbf{The optimized Phantom Twist.} (a) shows a static Phantom Twist and its components connected together using 0.8mm diameter carbon fiber rods. These components are (1) two batteries, (2) the control PCB, (3) the motor-propeller assembly, (4) a counterweight, and (5) the launch interface. (b,c) shows the long-exposure photo of the Phantom Twist flying taken with a 1/4\,s shutter speed, providing a coarse approximation of the Phantom Twist’s visual appearance to a human observer during flight. It is shown in flight against a white background (b) and against a more realistic environment (c). Supplementary video is available at \cite{vid}.}
    \label{fig:drone_image}
\end{figure}

Alternatively, rather than disappearing, some robots use biomimicry to appear as benign natural objects. Ornithopters and insect-scale flyers have been designed to mimic the visual profile and gait of birds or insects~\cite{chang2020soft,avhummingbird}, allowing them to hide in plain sight by appearing as part of the natural fauna~\cite{ward2015review}.

Other approaches have tried to use motion to hide a robot from an observer.  This approach termed ``motion camouflage''~\cite{srinivasan1995strategies, justh2006steering} has robots move in a way that hides their apparent motion to an observer~\cite{rano2016application}, reducing their chance of detection.

In contrast to these established methods, the potential of exploiting high-speed periodic motion to render robots invisible remains largely unexplored. Robots operating at high speeds or undergoing rapid periodic motion interact with human perception in fundamentally different ways. Because the human visual system has persistence of vision~\cite{wandell1995,efron1970minimum,hogben1974perceptual,ferry1892persistence}, fast motion can cause objects to appear blurred, translucent, or spatially diffuse rather than sharply defined. This temporal integration effect suggests an alternative path to reducing visual salience: rather than attempting to match background appearance or suppress motion cues, a robot can exploit high-speed motion and rotation to shape how it is perceived.

The surveillance drone Phantom Sentinel~\cite{risen2006boomerang} pioneered the application of this concept.  While limited information is available about the Phantom Sentinel, from available documentation, its rotational speed was insufficient to fully exploit visual persistence, resulting in a vehicle that is still largely visible. Beyond this specific attempt, other spinning platforms such as the Piccolissimo series drones~\cite{piccoli2017piccolissimo,curtis2023autonomous,wang2024single}, the Monospinner~\cite{zhang2016controllable, zhang2019design}, and various plant seed-inspired flyers~\cite{kim2021three, yang2024photochemically,win2019dynamics} inadvertently exhibit this blurring effect due to their flight dynamics. However, as their designs were not explicitly optimized for low visibility, they are either not spinning fast enough for the persistence effect to be apparent or possess high contrast and opaque parts after motion blur, preventing them from achieving effective reduction in visibility.

In this work, we introduce Phantom Twist, a single-propeller UAV designed to achieve low visibility by actively exploiting motion blur. By integrating a perceptual metric into the design loop, we optimize the spatial arrangement of functional components to minimize their visual footprint when spinning, rendering the UAV nearly imperceptible during flight.

\section{Phantom Twist Design and Optimization}

Phantom Twist uses a minimalist single-propeller UAV architecture similar to the Maneuverable Piccolissimo series UAVs~\cite{curtis2023autonomous, piccoli2017piccolissimo, wang2024single}. Unlike conventional multi-rotor UAVs, the Maneuverable Piccolissimo series drones and the Phantom Twist feature a single motor offset from the center of mass. As the motor and propeller spin, the rest of the UAV spins rapidly in the opposite direction, utilizing the gyroscopic effect of the rotating body to maintain passive stability. The UAV's position is actively controlled by varying the motor thrust throughout each rotation. 

The core components for Phantom Twist include a brushless DC motor with a propeller, two batteries, a main control PCB, and 0 to 2 counterweights (Fig.~\ref{fig:drone_image}). These components are structurally connected via thin carbon fiber rods. Due to the negligible mass and minimal visual cross-section of the structural rods, we exclude them from the design process. For simplicity, we will also ignore the effect of the launch interface and electrical wires, focusing the model on the primary functional components for computational efficiency.

Our overarching goal is to determine the optimal spatial configuration of these components that minimizes visual perceptibility while satisfying the dynamical constraints required for stable, controllable flight.

\subsection{Design Constraints}
\label{sec:design-constraints}

Based on related work on underactuated single-propeller UAVs~\cite{curtis2023autonomous, piccoli2017piccolissimo, wang2024single}, we identify ten specific constraints required to ensure Phantom Twist is physically realizable and capable of stable flight. These constraints fall into three general categories: mass and inertia, control and aerodynamics, and geometric. 

\noindent \textbf{Mass and Inertia Constraints}
\begin{enumerate}
    \item \textbf{Center of Mass:} Without loss of generality, we set the center of mass at the origin $\mathbf{O}$:
    \begin{equation}
        \vec{r}_{CoM} = \mathbf{O}
    \end{equation}

    \item \textbf{Principal Axis Alignment:} Due to their high spin rate, Phantom Twists' dynamics approximate those of rigid bodies under torque-free motion (motor torque $\times$ spin period $\ll$ angular momentum). In this regime, passive stability is only maintained when rotating about the axis of maximum or minimum inertia. In this work, we mandate rotation about the axis of maximum inertia. Without loss of generality, we set the axis of rotation to align with the Z-axis, which implies:
    \begin{equation}
        I_{xz} = I_{yz} = 0
    \end{equation}

    It is worth noting that because we point our motor along the negative Z axis ($\text{Z}^-$) for Phantom Twist, $\text{Z}^-$ will be close to the inverted gravity direction during flight.

    \item \textbf{Rotational Stability:} To enhance rotational stability, we further constrain the ratio $\eta=I_2/I_1$ between the first and second pricipal moment $I_1, I_2$. Physically, $\eta$ must lie between 0.5 and 1: values near 0.5 correspond to flatter geometries that remain highly visible from the side, while values near 1 make the rotational axis more sensitive to instantaneous thrust and other unmodeled perturbations, weakening the torque-free motion approximation used in constraint 2. We therefore use the balanced midpoint $\eta_0 = 0.75$ and require $\eta\leq\eta_0$. This constraint can be expanded using the elements of the inertia tensor:
    \begin{equation}
    \begin{split}
        (I_{xx} - \eta I_{zz})(I_{yy} - \eta I_{zz}) - I_{xy}^2 &\ge 0 \\
        \text{and} \quad 2 \eta I_{zz} - I_{xx} - I_{yy} &\ge 0
    \end{split}
    \end{equation}

    \item \textbf{Total Mass:} We selected a GEPRC GR1102 brushless DC motor with a maximum thrust of approximately 80\,g to drive our drone. Since the average thrust needs to equal the weight of the drone $m_{tot}$ when hovering and maximum control authority is achieved by pulsing thrust between $2m_{tot}$ and 0, we require the following to obtain maximum control authority:
    \begin{equation}
        m_{tot} \le 40\,\text{g}
    \end{equation}

    \item \textbf{Counterweight Feasibility:} All added counterweights must have non-negative mass:
    \begin{equation}
        m_{i} \ge 0
    \end{equation}
\end{enumerate}

\noindent \textbf{Control and Aerodynamic Constraints}
\begin{enumerate}[resume]
    \item \textbf{Control Authority:} Phantom Twist is controlled using the torque generated by the motor offset from the center of mass. To obtain enough control authority, the arm of thrust $r_{motor,xy}$ must be sufficiently large \edit{}{according to prior analysis of related single-motor UAVs~\cite{curtis2023autonomous,wang2024single}}:
    \begin{equation}
        \|r_{motor,xy}\| \ge 15 \text{ mm}
    \end{equation}

    \item \textbf{Spin Rate Range:} The spin rate must be maintained between approximately $15 \sim 25$ rotations per second (rps). Below 15\,rps, the rigid-body assumption in constraint 2 fails, often leading to crashes, and the motion blur effect weakens. Above 25\,rps, the inertia of the motor's rotor and the propeller prevents the motor from tracking the rapid power modulation required for control, making the UAV unstable.

    \item \textbf{Wake Clearance:} To prevent thrust loss, no components may be placed within the immediate vicinity of the motor-propeller assembly or propeller's downstream wake.
\end{enumerate}

\vspace{5mm}
\noindent \textbf{Geometric Constraints}
\begin{enumerate}[resume]
    \item \textbf{Compact Volume:} To keep the profile small, we artificially limit all components to be within a cylindrical boundary defined by radius $r_{xy} \le 70$\,mm and height $z \in [-55, 55]$\,mm.

    \item \textbf{Component Non-intersection:} Physical components must not overlap.
\end{enumerate}

The equilibrium spin rate used in constraint 7 is determined by the balance between the propeller drag torque $M_{prop}$ and the aerodynamic drag torque of the remaining parts $M_{drag}$. Since Phantom Twist designs are typically around 35\,g, we consider $M_{prop}$ as a constant. We model $M_{drag}$ as the sum of contributions from the motor, batteries, PCB, and spherical counterweights but ignore the contribution from other minor components. The drag due to these components is a function of spin rate and their position and orientation relative to the axis of rotation. 

To further simplify $M_{drag}$ calculation, we neglect the velocity gradients across individual components and the aerodynamic interference between components. We assume that each component acts as a static, isolated body in a uniform, infinite free stream with flow velocity $\vec{v}_{air}$ equal to the negative of the component's centroid velocity. Specifically, $\vec{v}_{air} = -\vec{\omega} \times \vec{r}_{xy}$, where $\vec{\omega}$ is the Phantom Twist's spin angular velocity, and $\vec{r}_{xy}$ is the vector from the rotation axis to the component's centroid.

We approximate the motor as a cylinder and the counterweights as spheres, and assume that they both have a drag coefficient of $C_{D,sph} = 0.47$~\cite{Hoerner1965}. Thus, the drag torque generated by the motor and counterweights can be approximated using

\begin{equation}
\begin{split}
    M_{drag,motor} &= \frac{1}{2} C_{D,sph}\,\rho A_{motor} \omega^2 \|\vec{r}_{motor,xy}\|^3\\
    M_{drag,weights} &= \frac{1}{2} C_{D,sph}\,\rho \sum_i(\frac{1}{4} \pi D_i^2) \omega^2 \|\vec{r}_{i,xy}\|^3
\end{split}
\end{equation}

\noindent where $\rho$ is the air density, $A_{motor}=D_{motor}H_{motor}$ where $D_{motor}, H_{motor}$ are diameter and height of the motor, $D_i$ is the diameter of the $i$-th counterweight, and the term $\|\vec{r}_{xy}\|$ denotes the distance between the centroid of each component to the rotation axis. 

Using CFD simulation, we obtained the following coarse approximation for drag on cuboid-shaped components including the PCB and the batteries.

\begin{figure}[h]
    \centering
    \includegraphics[width=.5\linewidth]{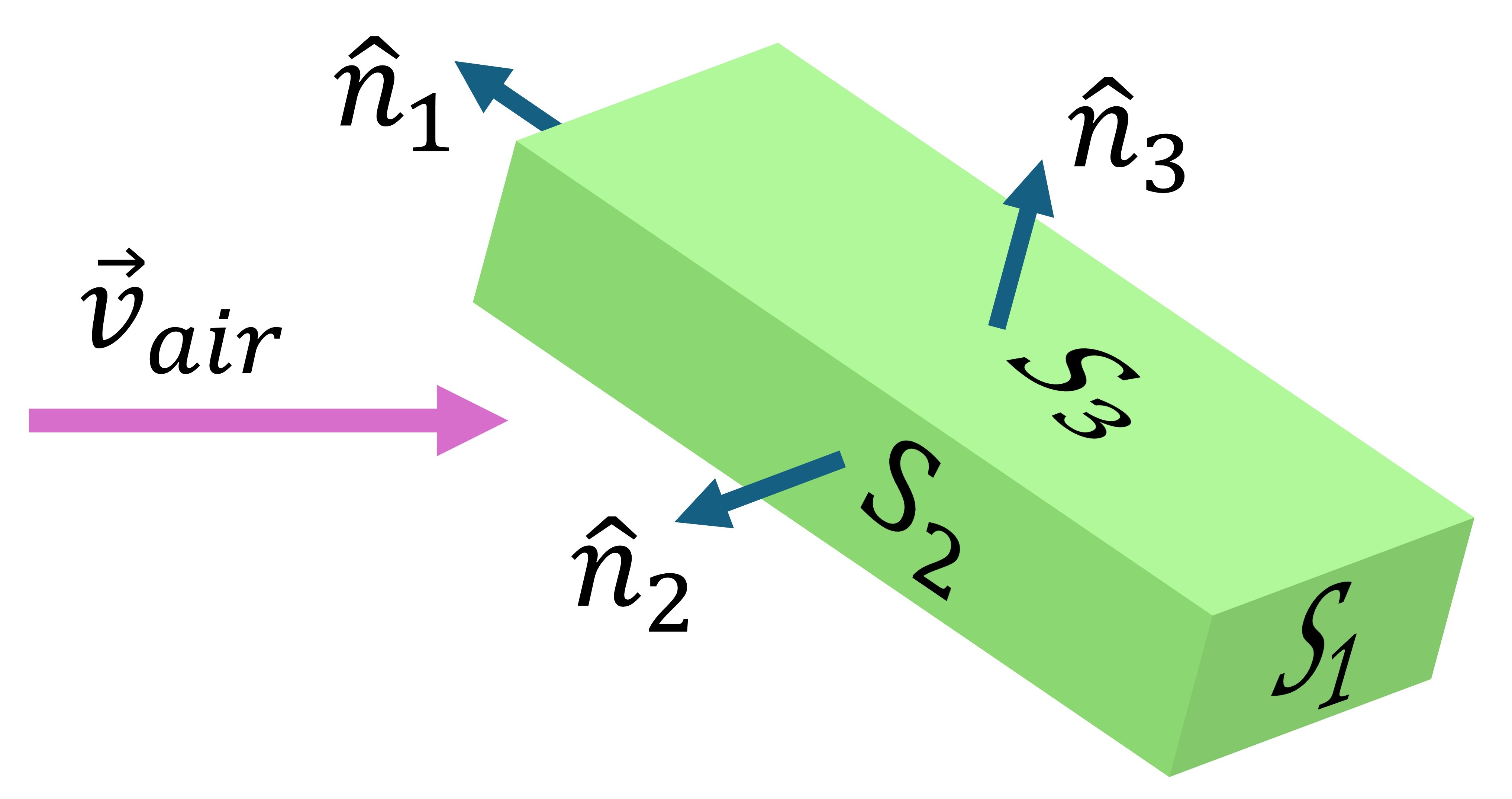}
    \caption{\textbf{Illustration for drag computation for cuboid-shaped components.} $S_i$ is the area of each surface of the cuboid, and $\hat{n}_i$ is the normal vector of each surface.}
    \label{fig:drag_illus}
\end{figure}

For a cuboid-shaped component shown in Fig.\ref{fig:drag_illus}, we can approximate its drag torque using:

\begin{equation}
\begin{aligned}
    M_{drag,cuboid} &= \frac{1}{2}\rho\sum_{i=1}^3 C_{1} S_i\,(\hat{n}_i\cdot\vec{v}_{air})^2 \|\vec{r}_{xy}\|\\
    &= \frac{1}{2}\rho\sum_{i=1}^3 C_{1} S_i\,(\hat{n}_i\cdot(\hat{\omega} \times \hat{r}_{xy}))^2 \omega^2\|\vec{r}_{xy}\|^3
\end{aligned}
\end{equation}


\noindent where $C_{1}=1.3$ is a dimensionless constant obtained from that CFD simulation, $\hat{\omega}, \hat{r}_{xy}$ are unit vectors in the direction of $\vec{\omega}, \vec{r}_{xy}$ respectively, and $S_i, \hat{n}_i$ are geometric properties of the cuboid as illustrated in Fig.\ref{fig:drag_illus}.

One can notice that the torque on all components except for the propeller is proportional to $\omega^2$, i.e. 

\begin{equation}
    M_{drag}=C_M \omega^2
\end{equation}

\noindent where $C_M$ is a drag torque coefficient determined by the geometric configuration of the Phantom Twist only. Thus, we can compute the spin angular velocity through

\begin{equation}
    \omega = \sqrt{\frac{M_{prop}}{C_M}}
    \label{eqn:CM}
\end{equation}

After building a few test drones and recording their spin rate during flight, we determined $M_{prop}$ in (\ref{eqn:CM}) and used (\ref{eqn:CM}) to further simplify constraint 7 to be a constraint about the geometric configuration of the Phantom Twist only:

\begin{enumerate}[label=7*)]
    \item \textbf{Spin Rate Range:} Drag torque coefficient $C_M$ is within $[20,60]\,\mathrm{g\cdot mm^2}$.
\end{enumerate}

As long as the designed Phantom Twist satisfies constraints 1-10 within reasonable manufacturing tolerances, we are confident that it can fly controllably.

\subsection{Optimization Objective: Visibility}

\begin{figure*}
    \centering
    \vspace{4mm}
    \includegraphics[width=1\linewidth]{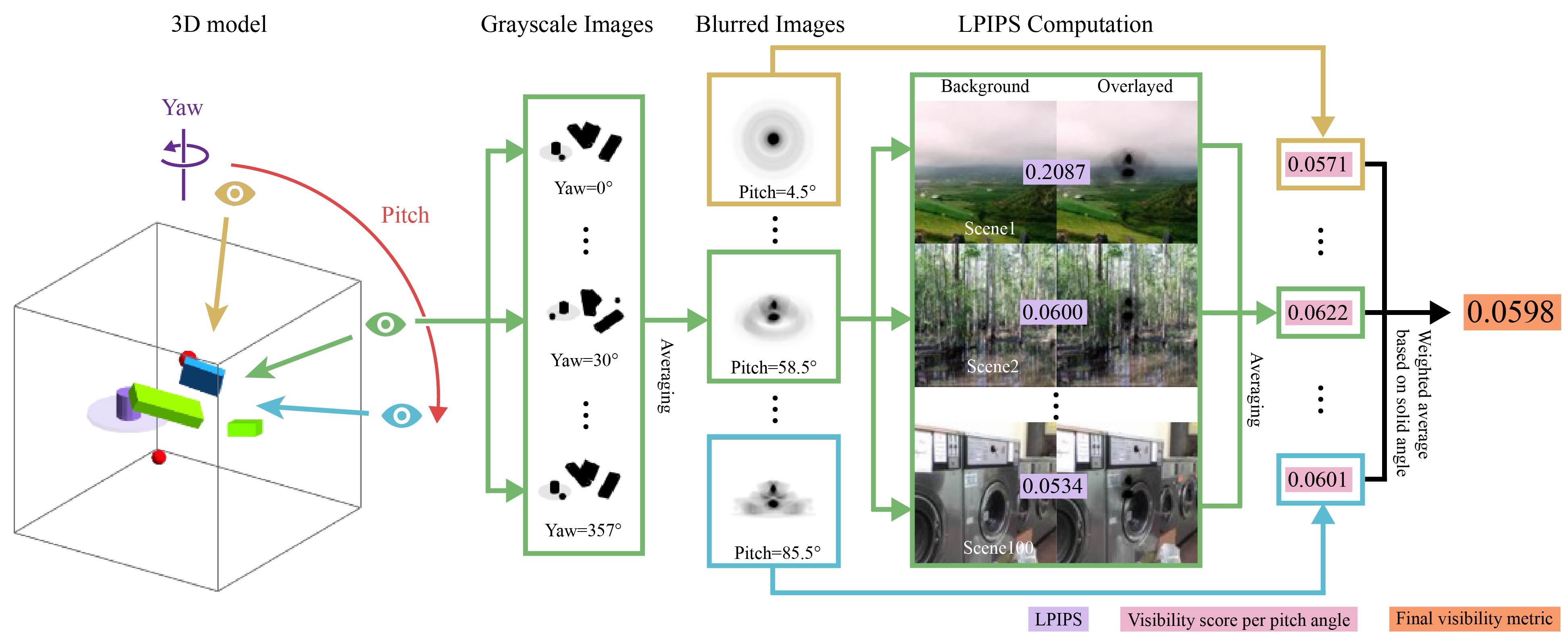}
    \vspace{1mm}
    \caption{\textbf{Computation of visibility metric.} The pipeline proceeds from left to right: First, for a pitch angle (here we use the pitch angle from the perspective of the green eye and arrow as an example), the 3D model (where batteries are green, PCBs are blue, motors and propellers are purple, and added counterweights are red) is rendered into 120 grayscale orthographic projections across a full yaw revolution. These images are averaged and gamma-corrected to synthesize a realistic motion-blurred appearance of the Phantom Twist for a specific pitch angle $\alpha$. This motion-blurred image is then superimposed onto 100 diverse backgrounds from the MiniPlaces dataset. We compute the LPIPS perceptual distance between each background and its overlayed version, averaging the results to obtain a visibility score for that pitch angle $\alpha$. Finally, we repeat this process to compute the scores from 10 distinct pitch angles and aggregate them via a weighted average (based on solid angle) to yield the final visibility metric. The equation for this process is given in (\ref{eqn:metric}). \edit{}{The red arrow indicates the pitch direction and the purple arrow indicates the yaw direction used in our rendering.}}
    \label{fig:vismetric}
    \vspace{3mm}
\end{figure*}

To quantify how visible a UAV is to a human observer, we first establish a model of its visual appearance based on its operating conditions. The overall procedure for visibility metric generation is shown in Fig.\ref{fig:vismetric}.

Given the miniature size of Phantom Twist (radius $\le 70$\,mm), perspective distortion becomes negligible even at relatively close viewing distances. This allows the observer's view to be approximated as an orthographic projection. Furthermore, because Phantom Twist operates at a spin rate of $15 \sim 25$\,rps comparable to the standard cinematic framerate of 24\,fps and consistent with the $50 \sim 150$\,ms persistence time of the human eye~\cite{efron1970minimum,hogben1974perceptual,wandell1995}, 
we assume the observer cannot resolve their instantaneous pose. 
In accordance with the Talbot-Plateau law~\cite{talbot1834xliv,plateau1830ueber,greene2023evaluating}, we model the UAV's appearance not as a static projection, but as the rotational average of its projection across all yaw angles, perceived as a motion-blurred, semi-transparent image.

To evaluate the perceptibility of this motion-blurred image, we utilize the Learned Perceptual Image Patch Similarity (LPIPS) metric~\cite{zhang2018LPIPS} which measures the difference between two images by extracting deep features from a pre-trained convolutional neural network (we used AlexNet) and computing the weighted Euclidean distance between activations. LPIPS has been widely adopted in computer vision to quantify perceptual differences, and has been shown to correlate better with human similarity judgments than simpler metrics (e.g., MSE, PSNR, SSIM)~\cite{zhang2018LPIPS} across a wide range of computer vision tasks~\cite{ding2021comparison}, image domains~\cite{long2025seeing,pettini2025synthesis}, and viewing conditions~\cite{chubarau2020perceptual}. We measure visibility by superimposing the motion-blurred image (gamma-encoded, with $\gamma$=2.2) of the Phantom Twist onto realistic background scenes and computing the LPIPS distance between the original and the superimposed images (Fig.\ref{fig:vismetric}), where a larger score indicates higher visibility. To ensure robustness, we average these scores over 100 diverse scenes sampled from the MiniPlaces dataset~\cite{zhou2017places} to compute the visibility score.

Since the observer's position relative to the Phantom Twist is unpredictable in a real-world deployment, we assume a uniform probability distribution for the viewing angle. The final optimization objective, the visibility metric, is therefore defined as the expectation of visibility score when the UAV is viewed from an arbitrary angle. Importantly, we assume that the robot is able to maintain a stable position relative to the observer, and leave for future work the increased visibility that could be caused by viewpoint changes if this assumption is violated.

Thus, our overall visibility metric takes the form of:

\begin{equation}
    \underbrace{\sum_{\alpha \in \mathcal{A}} \Omega(\alpha)}_{\circled{1}} \underbrace{\frac{1}{N_b} \sum_{b \in \mathcal{B}}}_{\circled{2}} \text{LPIPS}\left(b, \underbrace{\left(\frac{1}{N_{\phi}}\sum_{\phi\in\mathcal{P}}I(\alpha,\phi)\right)^{1/\gamma}}_{\circled{3}} b\right)
    \label{eqn:metric}
\end{equation}

\noindent where $\mathcal{A} = \{ \frac{\pi}{4N_\alpha}, \frac{3\pi}{4N_\alpha}, \dots, \frac{(2N_\alpha-1)\pi}{4N_\alpha} \}$ is the set of sampled pitch angles $\alpha$; we used $N_\alpha=10$ in our simulations. $\Omega(\alpha) = \cos(\alpha - \frac{\pi}{4N_\alpha}) - \cos(\alpha + \frac{\pi}{4N_\alpha})$ is the solid angle between $\alpha-\frac{\pi}{4N_\alpha}$ and $\alpha+\frac{\pi}{4N_\alpha}$. $\mathcal{B}$ is the set of $N_b=100$ background images from the MiniPlaces dataset \edit{}{and each $b\in\mathcal{B}$ denotes one specific background image}. $\mathcal{P} = \{ 0, \frac{2\pi}{N_\phi}, \dots, \frac{2(N_\phi-1)\pi}{N_\phi} \}$ represents the set of discrete yaw angles $\phi$ with $N_\phi=120$. $I(\alpha, \phi)$ denotes the rendered orthographic projection of the UAV in linear brightness space, and $\gamma=2.2$ is the gamma correction coefficient. The term labeled $\circled{1}$ represents the weighted average based on solid angle, the term labeled $\circled{2}$ represents the average over all background images, the term labeled $\circled{3}$ represents the synthesized motion-blurred UAV image, and the second argument of the LPIPS function represents the superposition of this result onto the background $b$.

The complete optimization problem can be written as

\vspace{-3mm}
\begin{equation}
    \mathop{\text{argmin}}\limits_{\mathbf{X}} \text{vis}(\mathbf{X}) \hspace{6mm} \text{s.t. constraints 1--10}    
    \label{eqn:opt_problem}
\end{equation}
\vspace{-3mm}

\edit{}{where $\mathbf{X}$ denotes the configuration of the drone that contains $23+4n$ variables for a design with $n$ counterweights (7 for the pose of each battery and PCB since we used angle-axis form to describe the orientation, 2 for the motor-propeller assembly's Y and Z position, 4 for the mass and position of each added counterweight), and $\mathrm{vis}(\mathbf{X})$ is the visibility of configuration $\mathbf{X}$ defined in (\ref{eqn:metric}).}

\subsection{Optimization Pipeline}

\begin{figure}[h]
    \centering
    \includegraphics[width=\linewidth]{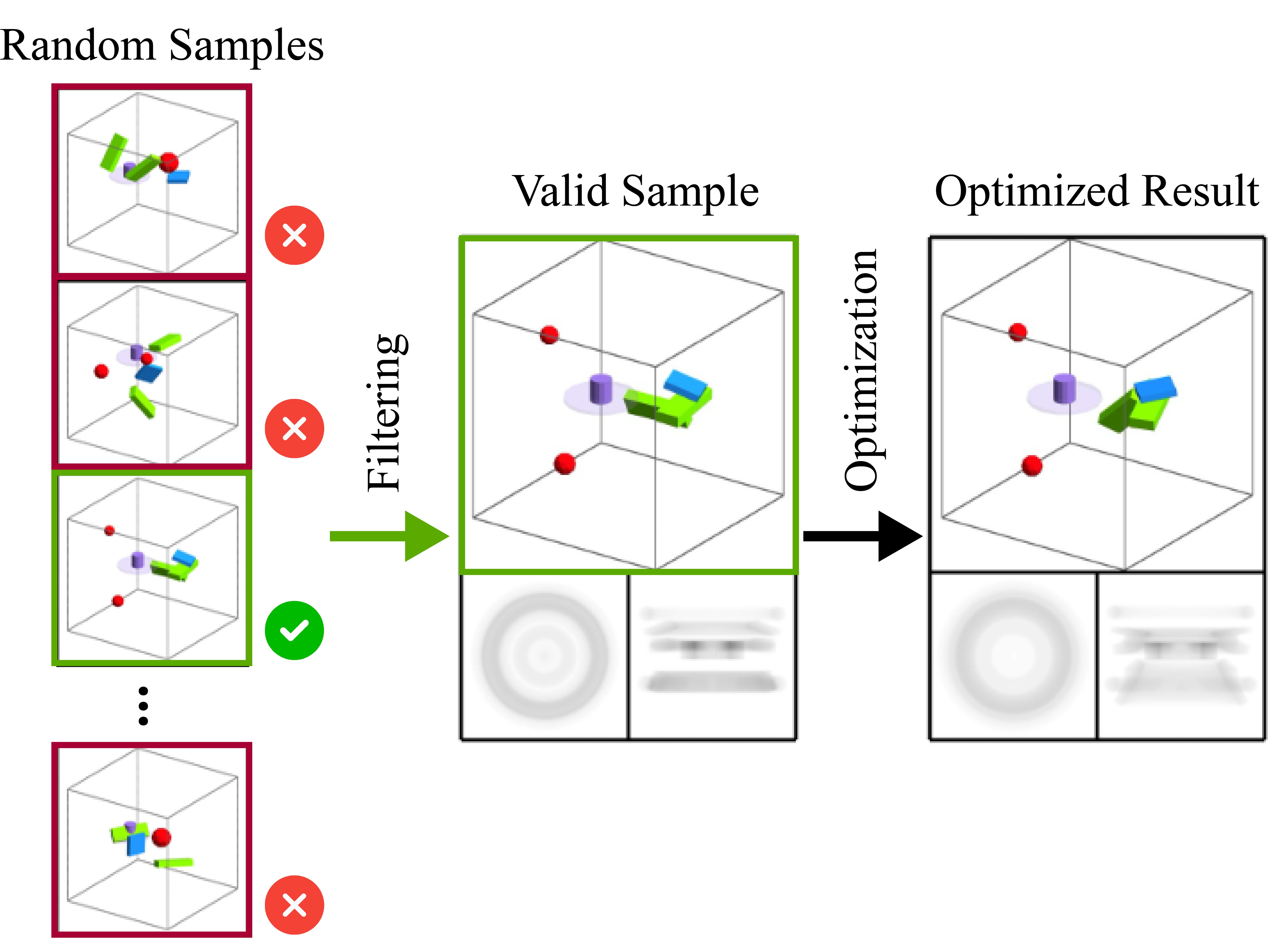}
    \caption{\textbf{Illustration of the two-stage optimization pipeline.} 
    The pipeline starts by randomly sampling component placements.
    The resulting designs were then passed through a feasibility filter that checks for constraint violations (see Sect.~\ref{sec:design-constraints}).
    A subset of the filtered designs then go through a gradient-based optimization process to further reduce their visibility. 
    In the example shown here, the visibility metric is reduced from 0.0167 to 0.0114 through optimization.
    Characteristic samples in each stage of the pipeline is shown in the figure. 
    The color coding for components is consistent with that in Fig.~\ref{fig:vismetric}. 
    Below the rendered 3D images are the motion-blurred top view (left) and side view (right) of the Phantom Twist.
    }
    \label{fig:pipeline}
\end{figure}

Our optimization process is divided into two stages: rapid feasibility filtering followed by gradient-based refinement. An illustration of this process is shown in Fig.\ref{fig:pipeline}.

We begin the search for valid initial designs in a first stage by randomly sampling the poses of the motor, batteries, and PCB. To improve computational efficiency, we apply a preliminary filter to discard samples that violate constraints 6, 8, 9, 10 prior to the addition of counterweights. For samples that pass this check, we use a numerical solver to determine the counterweight configurations (mass and position) with minimum total weight and satisfy the inertial constraints (1–5) and volume constraint (9). Finally, we verify these complete designs against the remaining constraints (6, 7, 8, 10) to ensure feasibility.

Starting from a feasible solution found in the first stage, we employ a gradient-based optimizer, the Sequential Least Squares Programming optimizer (SLSQP). This stage iteratively adjusts component poses and counterweight positions and masses to minimize the visibility metric while trying to strictly adhere to all constraints 1-10. 

\edit{}{The first stage counterweight configuration optimization is done through Mathematica's built-in NMinimize function, and the second stage optimization is done through the SLSQP solver in Python's SciPy package with a maximum iteration of 1000, $\text{ftol}=4.0 \times 10^{-6}$, and custom gradient computed using the central finite difference method, keeping all other parameters as default.}

\section{Phantom Twist Fabrication and Control}

%
%
%

To validate our computational design framework and verify the simulation results in the real world, we selected three distinct design configurations for physical fabrication and flight testing. The \textbf{human-designed baseline} is a manually designed configuration based on the Maneuverable Piccolissimo 3's geometry~\cite{wang2024single}; in this design, no additional counterweight is added and components are placed intuitively for balance and compactness, without explicit optimization for visual perceptibility. The \textbf{low-visibility sampled design} is a high-performing valid sample selected after the initial feasibility filtering stage but before second-stage optimization, representing a configuration that satisfies all flight constraints and offers low visibility but lacks fine-tuned gradient-based refinement. Finally, the \textbf{low-visibility optimized design} is the result of our complete two-stage optimization pipeline, representing the minimum visibility configuration found by our framework.


\subsection{Fabrication}

To build the Phantom Twist, we use a truss-based architecture to integrate the core components (motor, batteries, and PCB) as well as the extra components required for launch and flight.

\begin{figure}[h]
    \centering
    \includegraphics[width=\linewidth]{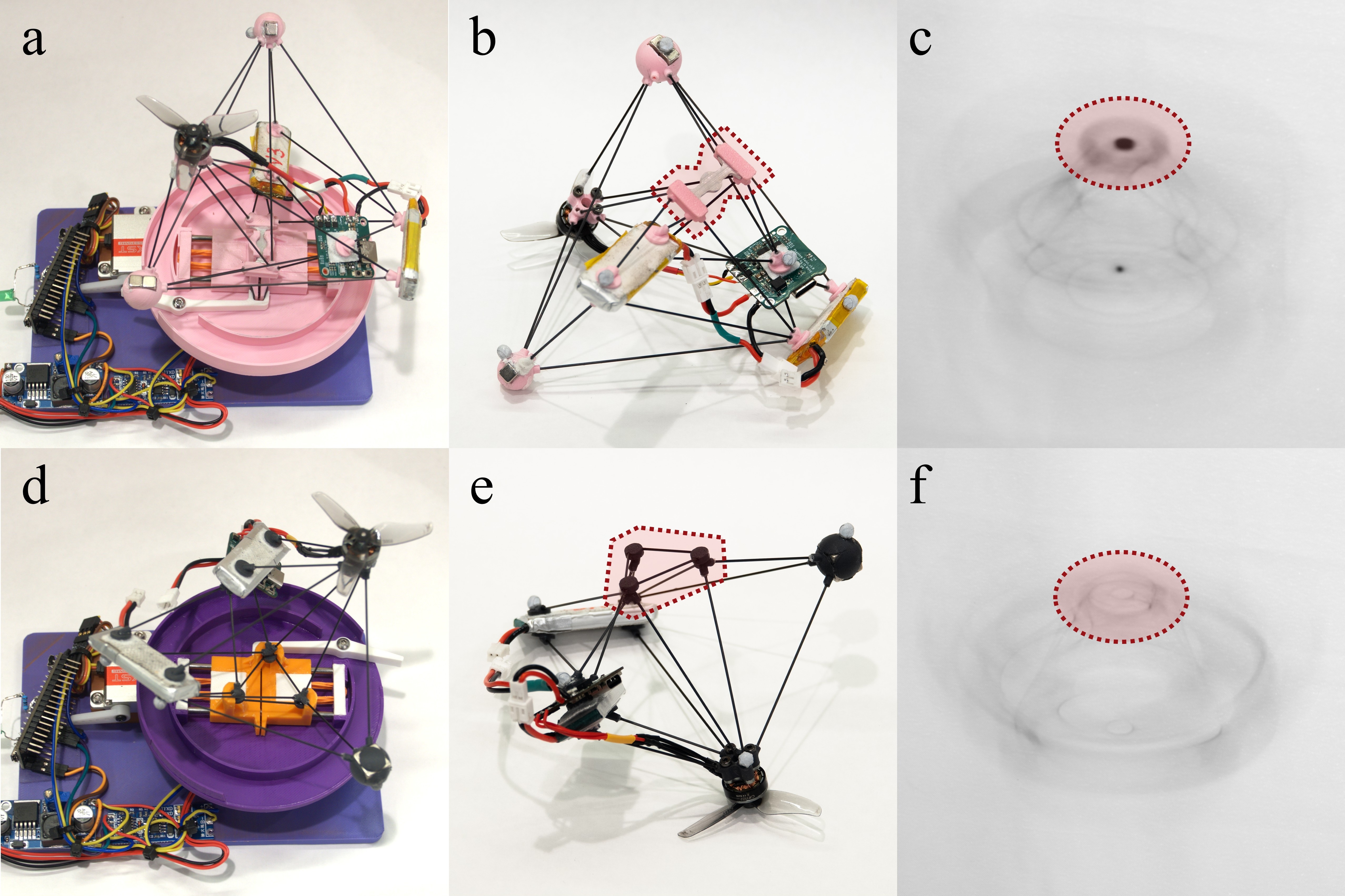}
    \caption{\textbf{Launch interface on Phantom Twists.} 
    Highlights of the launch interface's structure on the low-visibility sampled and optimized design. (a,d) shows the drone on the launcher, (b,e) shows the photo of static Phantom Twists with the launch interface highlighted in red, and (c,f) shows the long-exposure (1/4\,s) photos of them flying with the artifact created by the launch interface highlighted in red. (a,b,c) shows the unoptimized launch interface used on the low-visibility sampled design, and (d,e,f) shows the redesigned launch interface used on the low-visibility optimized design.} 
    \label{fig:launchinterface}
\end{figure}

In addition to the core components, the system also includes at most two counterweights, reflective tracking markers, and a launch interface. The counterweights are constructed by encapsulating dense steel blocks within 3D-printed PLA spherical shells to minimize volume while keeping their shape consistent with our assumptions. The launch interface, as shown in Fig.\ref{fig:launchinterface}.(b), is positioned directly above the center of mass. It enables an external launcher to spin the Phantom Twist to its stable operating spinning speed and then releases it, similar to the implementation described in~\cite{wang2024single}.

For the low-visibility optimized Phantom Twist, we extended our optimization process beyond component placement to refine the launching interface as well. After fabricating the human-designed baseline and low-visibility sampled prototypes, we observed that their standard monolithic launch interfaces created a high-contrast visual artifact in the long-exposure photos, as shown in Fig.\ref{fig:launchinterface}.(c). This is because these launch interfaces cross directly through the rotation axis, thus are not blurred out. To address this limitation, in the optimized design we developed a distributed truss structure (Fig.\ref{fig:launchinterface}.(e)) as the launch interface. As shown in Fig.\ref{fig:launchinterface}.(f), this re-designed launch interface ensures that no structural component is close to the rotating axis, preventing the formation of highly non-opaque features and further minimizing overall visibility.

All components are interconnected using $0.8$ mm carbon fiber rods. To securely attach the rods to the components, we designed a series of 3D-printed miniature mounting interfaces. The components and rods are all bonded to these interfaces using cyanoacrylate adhesive, creating a rigid, unified spatial structure.

This fabrication strategy ensures high structural stiffness to resist high centrifugal loads introduced by the fast spinning. Furthermore, these thin connecting rods have negligible mass and produce minimal aerodynamic drag, validating our modeling assumption that the structural framework can be neglected in the modeling process.

\begin{figure*}[htbp]
    \centering
    \includegraphics[width=\textwidth]{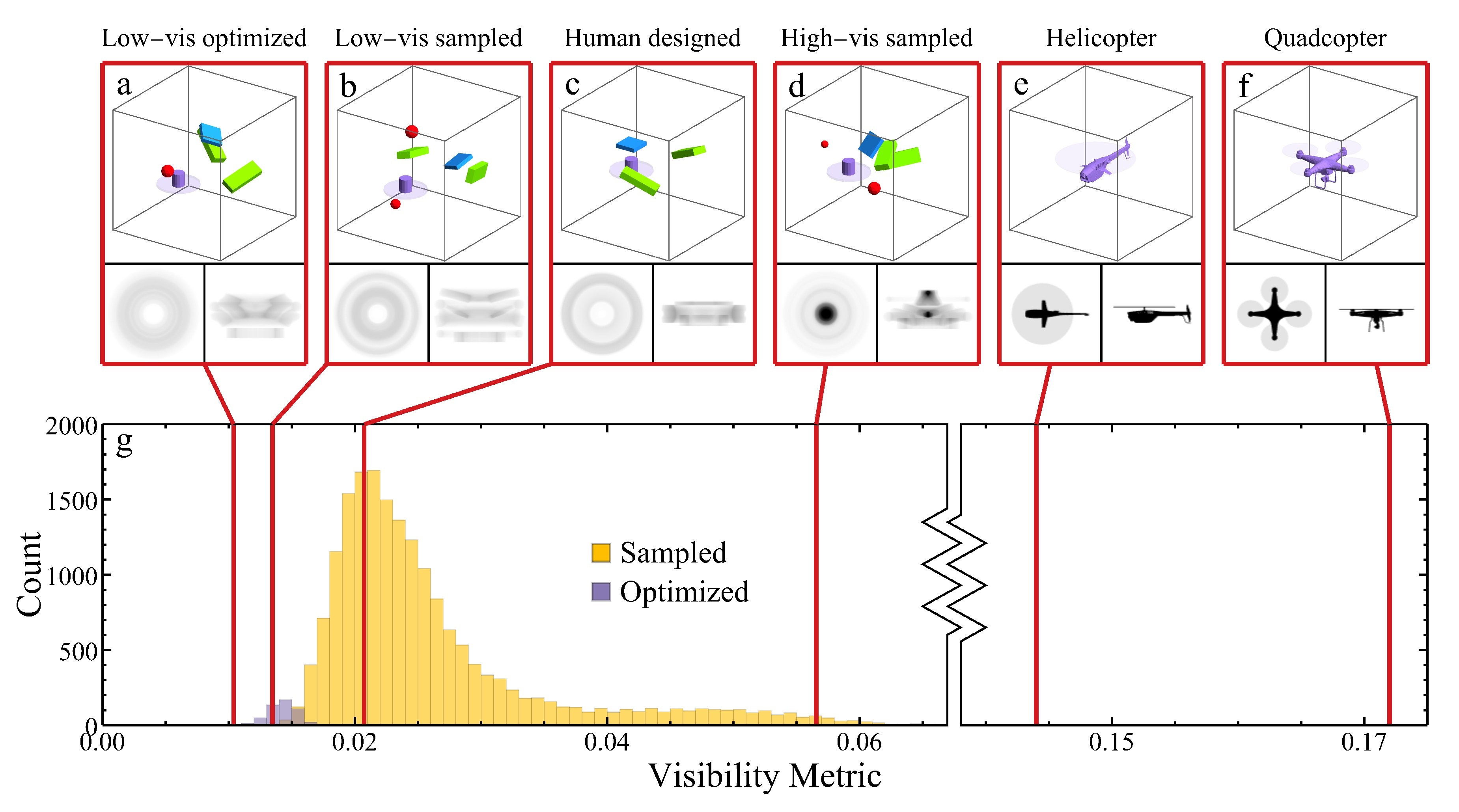}
    \caption{\textbf{The Phantom Twist design space and its visual appearance.} 
    (a-d) shows representative Phantom Twist configurations sampled from distinct performance regions. For each sample, we show the static structural configuration and the simulated motion-blurred appearance from top and side views. The color coding for components is consistent with that in Fig.\ref{fig:vismetric}.
    (e,f) shows baseline for helicopter and quadcopter as well as their top and side views.
    (g) is the histogram of visibility metric of all valid and optimized solutions found.
    The yellow samples represent the feasible configurations found via random sampling and filtering. 
    The purple samples show the best 500 designs after gradient-based refinement, demonstrating a significant reduction in the visibility metric.
    \edit{}{The zigzag symbol in the figure denotes a break in the axis.}
    }
    \label{fig:histogram_composite}
\end{figure*}

\subsection{Control}

As Phantom Twist flies, it spins at  $15 \sim 25$\,rps around its principal axis. The on-board controller (Espressif ESP32-S3-PICO) executes a single-actuator control law similar to~\cite{curtis2023autonomous,wang2024single} to control the drone. The motor executes two power commands per rotation: the average of commands regulates vertical thrust, while the difference of commands and the switching time regulate the average torque that tilts the Phantom Twist and in turn generates horizontal acceleration. In this work, an external OptiTrack system is used for state estimation.

As this control strategy relies on applying thrust at a specific time and UAV pose, synchronization between state estimation and actuation is critical: a timing error exceeding 5ms can cause loss of control. To address this, we deploy an ESP32-S3 router to capture precise frame timestamps directly from the OptiTrack e-sync device. These timestamps are aggregated with the filtered tracking data from the computer and transmitted to the Phantom Twist via low-latency ESP-NOW protocol. Using this timing reference, a custom protocol synchronizes the Phantom Twist’s internal clock with the Optitrack system to within 0.1ms, effectively rejecting network jitter and ensuring that motor commands are executed with precise timing.


\section{Results}

\begin{figure*}[h]
    \centering
    \includegraphics[width=\textwidth]{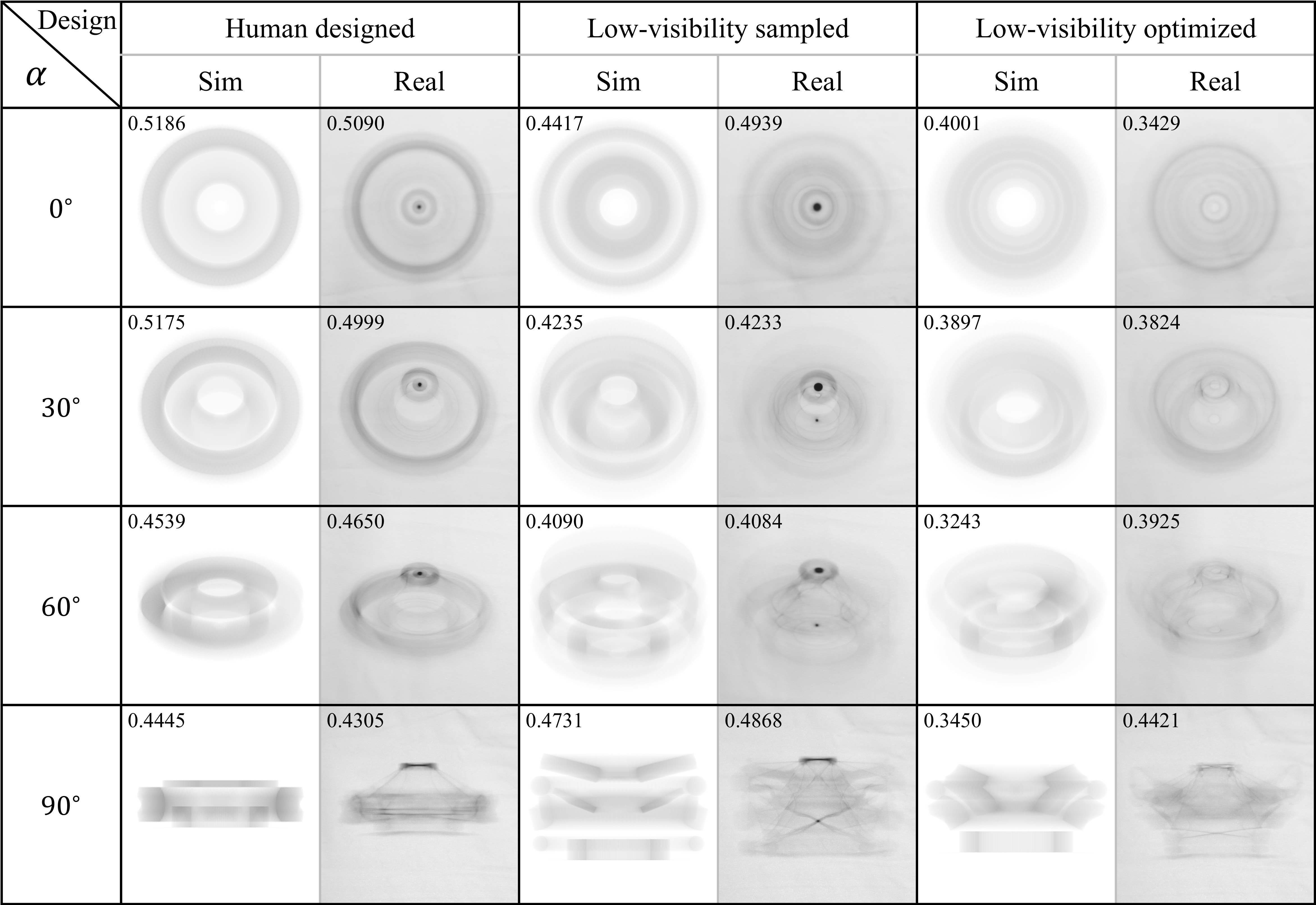}
    \caption{\textbf{Simulated visual versus long exposure photos.} 
    We compare the visuals of the three fabricated Phantom Twist prototypes: the human designed baseline (left two columns), the low-visibility sampled Phantom Twist (middle two columns), and the low-visibility optimized Phantom Twist (right two columns). For each design, the simulated visual prediction is presented alongside the corresponding real-world long-exposure photo. The rows illustrate these comparisons across varying viewing angles, ranging from a top-down view ($\alpha=0^\circ$) to a side view ($\alpha=90^\circ$). Note the strong correspondence between the simulated images and the real-world photos across all viewing angles. We adjusted the exposure of the long-exposure photos using Adobe Lightroom so the background intensity is consistent across photos. The number on the left upper corner of each image is the LPIPS score between that image and its corresponding background.
    }
    \label{fig:sim_vs_real}
    \vspace{5mm}
\end{figure*}

\subsection{Simulation and Optimization Analysis}
Through feasibility filtering, we generated a dataset of around 20,000 feasible Phantom Twist configurations. These configurations satisfy all constraints 1-10 but vary in component arrangement. Among these configurations, we picked the 500 lowest visibility configurations and further optimized them in the gradient-based refinement step.

Figure \ref{fig:histogram_composite} provides a comprehensive overview of the results. The quadcopter and helicopter shown in Fig.\ref{fig:histogram_composite}.(e,f) have an order of magnitude higher visibility metric compared to the best Phantom Twist designs as they contain large, opaque areas with sharp edges. Furthermore, as shown in Fig.\ref{fig:histogram_composite}.(a-d), when the LPIPS score decreases, the motion-blurred image of the Phantom Twist transforms from a more solid object into a faint, translucent volume with minimal high-contrast features. As shown in Fig.\ref{fig:histogram_composite}.(d), high-visibility designs from the feasible set often suffer from completely non-transparent areas created by components intersecting the rotation axis or dark areas created when rotational envelopes of multiple components overlap and superimpose, compounding their opacity. The human-designed baseline also suffers from these issues, thus having a metric of 0.0207, confirming that intuitive component placement similar to MP3 in~\cite{wang2024single} is not sufficient for minimizing visibility. In contrast, the low-visibility sampled design shown in Fig.\ref{fig:histogram_composite}.(b) has much lower visibility metric of 0.0135 as a result of placing components further from the center of rotation and distributing them across different heights. The optimization process even further reduced the minimum visibility metric to 0.0104 by fine-tuning the position and orientation of objects, especially reducing features like high-contrast rings or sharp edges, as shown in Fig.\ref{fig:histogram_composite}.(a). 

For the 500 best samples that are optimized, the average visibility metric is 0.0161 with a minimum of 0.0134 and a maximum of 0.0169 before optimization. After optimization, the average visibility metric is reduced to 0.0142 with the minimum and maximum of 0.0104 and 0.0169. On average, the optimization process reduced the visibility metric by 12\%.

\subsection{Physical Validation}

To validate the simulation results, we fabricated the three characteristic designs for comparative analysis.

When the drone is flying, we captured long-exposure photos (1/4\,s shutter time, F16.0 equivalent aperture, ISO \,64, 135\,mm equivalent focal length) and 30x slow-motion videos using Sony's RX10-M4 camera from approximately 1.8m away at $0^\circ, 30^\circ, 60^\circ, \text{and } 90^\circ$ inclination. These long-exposure photos serve as a coarse approximation of the Phantom Twist's visual appearance to a human observer during flight.

A critical concern with asymmetric single-propeller UAVs is flight stability. Despite the counter-intuitive placement of the components, all three Phantom Twists can maintain stable hover for approximately 10 minutes with a RMS position error of around 3\,cm and a max position error less than 10\,cm. The flight trajectory is similar to that in related work \cite{curtis2023autonomous, wang2024single}. This suggests that our constraints successfully capture the sufficient conditions for stable flight.

The comparison between the Phantom Twist's simulated appearances and their long-exposure flight photos is shown in Fig.\ref{fig:sim_vs_real}. As shown in the figure, the simulation accurately predicts the visual features in the real-world photos except for the unmodeled rods, wires, and launch interface. While the human-designed baseline creates distinct, high-contrast concentric rings, the low-visibility samples and the optimized sample are much less visible. Within this low-visibility regime, the optimized design exhibits distinct structural differences: the adjusted arrangement of core components results in reduced apparent dark bands around $\alpha=0^\circ$ and a smaller overall projected area around $\alpha = 90^\circ$. Furthermore, the structural optimization of the launch interface eliminates the dark feature near the rotation axis which can be seen in long-exposure photos of the baseline and low-visibility sampled designs.

\begin{figure}[htbp]
\vspace{5mm}
    \centering
    \includegraphics[width=1\linewidth]{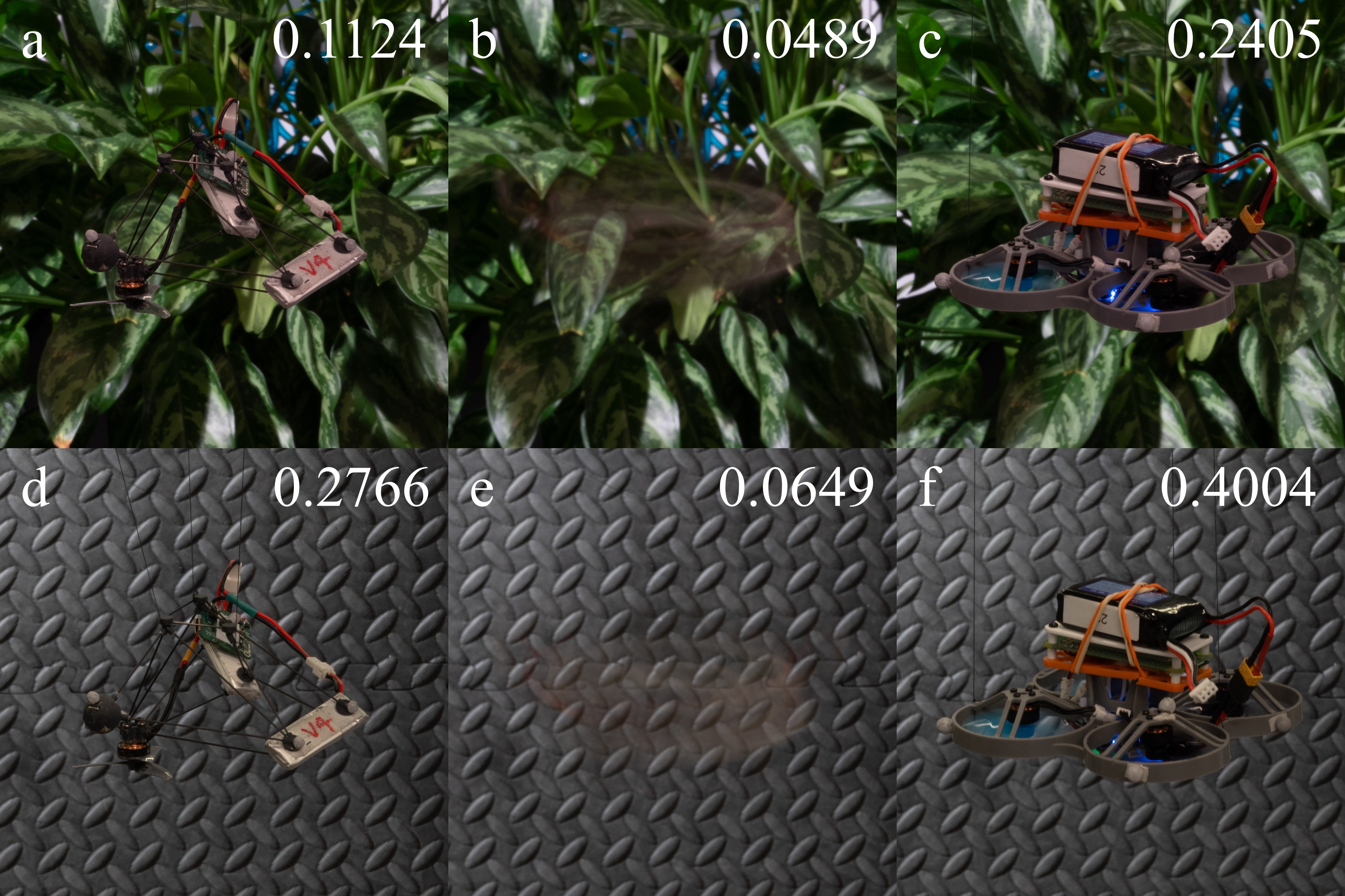}
    \caption{\textbf{Comparison of visibility.} Three columns show static optimized Phantom Twist (a,d), dynamic hovering optimized Phantom Twist (b,e), and a static quadcopter in front of the same background (c,f). Images in the same row are captured with the same background and the same camera parameters. \edit{}{The number on the right upper corner of each image is the LPIPS score between that image and its corresponding background.}}
    \label{fig:quadcompare}
    \vspace{5mm}
\end{figure}

We also compared our optimized Phantom Twist with a conventional quadcopter in a more realistic setting in Fig.\ref{fig:quadcompare}. The static Phantom Twist or the quadcopter is significantly more visible than the motion-blurred Phantom Twist.

\vspace{4mm}
\section{Discussion}
\label{sec:discussion}

In this work, we approximated the visual appearance of the spinning UAV as a static, motion-blurred image, relying on the assumption that the spin rate ($15 \sim 25$\,rps) is high enough to fully blur the Phantom Twist's appearance. However, 
human sensitivity to flicker has been reported to frequencies as high as 500\,Hz in some settings~\cite{davis2015humans}, with standard flicker perception varying with luminance according to the Ferry-Porter law~\cite{kremers2024erg,ferry1892persistence,porter1902contributions}. This law can be used to predict flicker perceptibility under varying illumination conditions, and explains the observation of some flicker when Phantom Twist is brightly lit. \edit{}{Nevertheless, much like the 24\,fps convention in cinematic imaging, the 15--25\,rps operating range produces limited flicker in most of our tested lighting conditions, and during experiments the vehicle remained predominantly motion blurred.} Similarly, spatial drift or oscillation caused by imperfect control can also make the Phantom Twist more visible. Consequently, our simplified model may underestimate visibility in dynamic real-world conditions. Future iterations of this pipeline could incorporate additional perceptual metrics to account for these dynamic effects, and future iterations of the drone can also use new hardware that allows for faster rotations, reducing the flicker.

Another limitation is from the non-idealities of the physical fabrication. Our optimization framework focused exclusively on the placement of core components. However, connecting rods, wiring, and the launch interface remain discernible in long-exposure photographs, despite being significantly less conspicuous after optimization. Future designs could take these minor components into consideration in the design pipeline or use more optically transparent materials (i.e. clear plastic rods,  transparent PCBs) for these components to further reduce visibility.

While a fully rotating UAV such as the Phantom Twist may initially appear challenging to use in practical settings, its continuous rotation can be leveraged as a distinct advantage. In fact, similar platforms in \cite{wang2024single} use this rotation to their advantage and achieved $360^\circ$ communication and relative localization using simple infrared transceivers. Other additions to the UAV such as LEDs, time-of-flight (ToF) sensors, or line-scan cameras would allow for tasks like $360^\circ$ imaging or in-the-air display \cite{cai2025quadrotary}.

This work focuses on reducing visual detectability, however acoustic emissions are also a critical factor influencing the perceptibility of a UAV. Propeller sound can significantly affect both human awareness and interaction, particularly in close-range or quiet environments. Addressing acoustic signatures, however, lies beyond the scope of this paper. Future efforts toward low-observability of UAVs should integrate visual minimization with quiet propulsion strategies, such as using propeller designs optimized for reduced acoustic output~\cite{quietprop} or other possible propulsion concepts still under active development like electrohydrodynamic thrust~\cite{EHD_thrust}.

\section{Conclusion} 
\label{sec:conclusion}

In this work, we presented Phantom Twist, a type of single-propeller UAV designed to achieve low visibility by exploiting the motion blur caused by its high-speed rotation. We identified the sufficient inertial and aerodynamic constraints for flight stability and proposed a computational method to evaluate the UAV's visibility using LPIPS. Building on these, we developed a two-stage computational design pipeline that optimizes the spatial arrangement of the main components to minimize visual perceptibility while ensuring that the resulting single-propeller UAV can fly stably.

We validated our approach through the fabrication and flight testing of three distinct prototypes: a human-designed baseline, a high-performing random sample, and the fully optimized design. Flight experiments confirm that our constraints successfully capture the sufficient conditions for stable flight, with all prototypes capable of controlled hovering. In flight tests, we observed that the optimized UAV exhibits significantly reduced visual perceptibility compared to a conventional quadcopter.

This work establishes a novel design paradigm for low-visibility UAVs, moving beyond traditional static intensity or pattern-based concealment to leverage the persistence effect.

\section*{Acknowledgment}

David Matthews was supported by NSF GRFP, grant number DGE-2234667.

\bibliography{references}

\end{document}